\documentclass[lettersize,journal]{IEEEtran}
\usepackage{amsmath,amsfonts}
\usepackage{algorithmic}
\usepackage{algorithm}
\usepackage{array}
\usepackage{textcomp}
\usepackage{stfloats}
\usepackage{url}
\usepackage{verbatim}
\usepackage{graphicx}
\usepackage{url}
\usepackage{graphicx}
\usepackage{booktabs}
\newcommand{\mynocheck}{\textcolor{textgray}{$\times$}}
\newenvironment{tightcenter}{%
  \setlength\topsep{3pt}
  \setlength\parskip{3pt}
  \begin{center}
  \begin{minipage}{.4\textwidth}
}{%
  \end{minipage}
  \end{center}
}

\usepackage{wrapfig}
\usepackage{makecell}
\usepackage{multirow}
\usepackage{subcaption}
\usepackage[table]{xcolor}
\usepackage{xcolor}         
\usepackage{bm}
\usepackage{lineno}
\usepackage[colorlinks,linkcolor=blue,citecolor=blue]{hyperref}

\definecolor{mygray}{gray}{.9}
\definecolor{rouse}{rgb}{0.981,0.961,0.941}
\newcommand{\second}[1]{{\color{blue}{\textbf{#1}}}}
\definecolor{rouse}{rgb}{0.981,0.961,0.941}
\definecolor{light-yellow}{rgb}{1,1,0.93}
\definecolor{light-green}{rgb}{0.95,1,0.95}
\definecolor{light-blue}{rgb}{0.95,0.95,1}
\definecolor{textgray}{rgb}{0.5,0.5,0.5}
\definecolor{defaultColor}{RGB}{230, 230, 250}

\begin{document}
\IEEEpubidadjcol

\title{UniQA: Unified Vision-Language Pre-training for Image Quality and Aesthetic Assessment}

\author{Hantao Zhou, Longxiang Tang, Rui Yang, Guanyi Qin, Yan Zhang, Yutao Liu, Xiu Li, Runze Hu, Guangtao~Zhai,~\IEEEmembership{Fellow, IEEE}
\thanks{Hantao Zhou, Longxiang Tang, Rui Yang, Guanyi Qin, Xiu Li are with Tsinghua Shenzhen International Graduate School, Tsinghua University, ShenZhen, 518055, China (e-mail: hantaozh@outlook.com, lloong.x@gmail.com,
r-yang20@mails.tsinghua.edu.cn,
qgy21@mails.tsinghua.edu.cn,
li.xiu@sz.ts
inghua.edu.cn).

Runze Hu is with the School of Information and Electronics, Beijing Insti- tute of Technology, Beijing 100086, China (e-mail: hrzlpk2015@gmail.com).

Yan Zhang is with Media Analytics and Computing Lab, Department of Artificial Intelligence, School of Informatics, Xiamen University, 361005, China (e-mail: bzhy986@gmail.com)

Yutao Liu is with the School of Computer Science and Technology, Ocean University of China, Qingdao 266100, China (e-mail: liuyutao@ouc.edu.cn)

Guangtao Zhai is with the Institute of Image Communication and Information
Processing, Shanghai Jiao Tong University, Shanghai 200240, China (e-mail:
zhaiguangtao@gmail.com).

}}

\markboth{JOURNAL OF \LaTeX\ CLASS FILES}%
{UniQA: Unified Vision-Language Pre-training for Image Quality and
Aesthetic Assessment}

\IEEEpubid{0000--0000/00\$00.00~\copyright~2021 IEEE}

\maketitle

\begin{abstract}
Image Quality Assessment (IQA) and Image Aesthetic Assessment (IAA) aim to simulate human subjective perception of image visual quality and aesthetic appeal.
Despite distinct learning objectives, they have underlying interconnectedness due to consistent human assessment perception.
In this paper, we propose \textbf{Uni}fied vision-language pre-training of \textbf{Q}uality and \textbf{A}esthetics (\textbf{UniQA}), to extract useful and common representations from two tasks, thereby benefiting them simultaneously.
However, the lack of text in the IQA datasets and the textual noise in the IAA datasets pose severe challenges for multimodal pre-training. To address this, we (1) utilize multimodal large language models (MLLMs) to generate high-quality text descriptions; (2) use the generated text for IAA  as metadata to purify noisy IAA data.
To effectively adapt the pre-trained UniQA to downstream tasks, we further propose a lightweight adapter that utilizes versatile cues to fully exploit the extensive knowledge of the pre-trained model. 
UniQA demonstrates high competitiveness in various image assessment tasks, including classical IQA and IAA tasks, few-label IQA, and other downstream tasks, showing promise as a foundational assessment model. Codes are available at
\href{https://github.com/zht8506/UniQA}{https://github.com/zht8506/UniQA}.
\end{abstract}


\section{Introduction}
Image Quality Assessment (IQA)\footnote{The IQA in this work refers to the no-reference IQA.} and Image Aesthetic Assessment (IAA) aim to measure the perceived quality and beauty of an image. They find broad applications in many scenarios, such as guiding individuals in image photography and editing, and serving as tools for image dehazing model~\cite{zhao2021refinednet}. Consequently, huge efforts~\cite{su2020blindly,ke2021musiq,he2022rethinking,ke2023vila} have been devoted to establishing effective IQA and IAA models.

IQA and IAA concentrate on distinct aspects of image assessment, with IQA primarily focusing on the distortion level of the image, while IAA is oriented towards evaluating the aesthetic appeal of the image. 
Despite their differences, IQA and IAA have underlying commonality: \textbf{simulating human subjective perceptions of images.}
Specifically, in human subjective image evaluation, quality and aesthetics exhibit a mutual influence~\cite{gotz2023inter}, 
such that high-quality images are more likely to possess a higher aesthetic appeal compared to their low-quality counterparts.
Thus, the learning process for both tasks not only acquires features unique to themselves but also involves the learning of task-agnostic common representations~\cite{jenadeleh2017blind}. This commonality sparks an idea:
\begin{tightcenter}
    \textit{Can we develop a foundational model with robust visual assessment perceptions consistent with human to benefit both IQA and IAA tasks?}
\end{tightcenter}

Several existing works~\cite{ke2021musiq,zhang2023synergetic,sheng2023technical} have explored the relationship between IQA and IAA tasks from different perspectives. For instance, some works (\textit{e.g.}, MUSIQ~\cite{ke2021musiq}) can be applied to IQA and IAA tasks indiscriminately, but they cannot exploit beneficial representations from another task. 
Q-Align~\cite{wu2023q} and DSINet~\cite{wei2024dual} also find the similarities of two tasks and attempt to tackle them with unified architecture and training. However, they typically unify datasets of two tasks for regression training directly, which cannot explicitly learn the task-shared representations, restricting the extraction of mutual benefits. TQ4AQ~\cite{sheng2023technical} develops a quality-assisted image aesthetic quality assessment method that utilizes quality information to improve the IAA task, but only considers the impact of IQA on IAA. In this paper, we propose the \textbf{Uni}fied pre-training of \textbf{Q}uality and \textbf{A}esthetics (UniQA) to extract mutually beneficial and effective representations for both tasks. Then, the pre-trained UniQA can be flexibly applied to IQA and IAA datasets.

\begin{figure*}[t]
\centering
\includegraphics[scale=0.57]{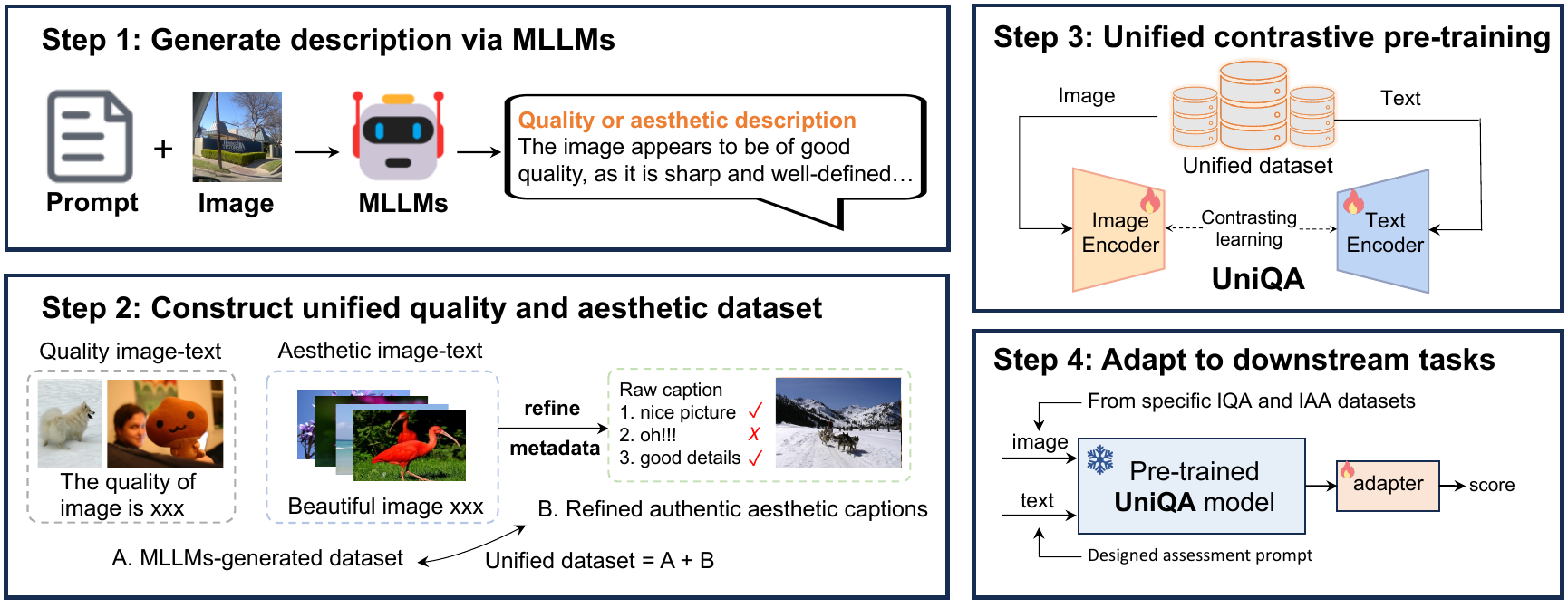}
\caption{The overview of our method. We leverage MLLMs to generate quality- and aesthetics-related descriptions (Step 1) and utilize the generated data to refine authentic noisy data (Step 2). We conduct unified pre-training to obtain UniQA (Step 3), which can be flexibly applied to both IQA and IAA tasks with a lightweight adapter (Step 4).
}
\label{fig:intro}
\end{figure*}

To achieve unified pre-training, a straightforward solution involves consolidating all IQA and IAA datasets and then training the model to regress towards the mean opinion scores (MOS) annotated by humans. 
However, existing datasets show variations in perceptual scales due to differences in subjective testing methodologies~\cite{zhang2021uncertainty}. 
As a result, this training strategy makes the model develop a score bias toward larger scale datasets.
Moreover, it may not effectively capture the unique characteristics of IQA and IAA, as the MOS labels cannot be explicitly interpreted. 
To this end, we propose to use {\textbf{text descriptions}} 
as a bridge to integrate the two tasks, leveraging the rich and fine-grained semantics inherent in text to provide more auxiliary information.

\IEEEpubidadjcol

To achieve unified pre-training, we need to construct image-text data for IQA and IAA tasks. However, existing IQA datasets~\cite{live,clive,koniq,spaq,kadid,ying2020patches} typically have images only and lack text descriptions. 
While IAA dataset~\cite{avacaption} include text data provided by humans, they often contain considerable textual noise irrelevant for aesthetic assessment.
Therefore, a top priority is determining how to acquire high-quality image-text data for both tasks. Recently, multimodal large language models (MLLMs)~\cite{bai2023qwen,liu2024visual,lin2023sphinx,li2023otter} have demonstrated outstanding capabilities in image understanding, which can generate reasonable responses based on images and user instructions. Inspired by this, we propose utilizing MLLMs with tailored prompts to generate  quality- and aesthetics-related  descriptions for the IQA and IAA datasets (Step 1 of Fig.~\ref{fig:intro}). As observed in Fig.~\ref{fig:prompt}, this approach provides a comprehensive and precise depiction of image quality and aesthetics. Furthermore, we utilize these generated high-quality aesthetic  descriptions as metadata to refine the raw aesthetic caption dataset (Step 2 of Fig.~\ref{fig:intro}). Finally, we unify the IQA and IAA datasets to conduct multimodal pre-training (Step 3 of Fig.~\ref{fig:intro}) to obtain UniQA.

To effectively adapt the pre-trained
UniQA to downstream tasks, we propose a lightweight adapter, namely the Multi-Cue Integration Adapter (Step 4 of Fig.~\ref{fig:intro}). 
This adapter uses versatile cues related to image assessment to prompt the pre-trained UniQA, adeptly extracting useful knowledge and comprehensively assessing the image. 
With much fewer tunable parameters compared to previous IQA and IAA models, our model outperforms them on both tasks.
More surprisingly, our method achieves impressive results on few-label IQA, \textit{e.g.}, achieving the SRCC values of 0.828 (vs. 0.760 on CLIVE of SOTA method GRepQ~\cite{srinath2024learning}). UniQA also generalizes well to various downstream image assessment tasks, highlighting its strong foundation and generalization capabilities.

Our contributions can be summarized as follows:
\begin{itemize}
    \item With the assistance of MLLMs, we construct a high-quality image-text dataset about image quality and aesthetics. 
    Thro-ugh pre-training on this dataset, we develop UniQA, which effectively learns a general perception of image assessment, promoting the effective and efficient learning of both IQA and IAA tasks.
    \item 
    We propose a novel Multi-Cue Integration Adapter, which integrates various assessment-related cues to fully exploit the extensive knowledge of the pre-trained model with minimal additional parameters.
    \item 
    Extensive experiments show that our method achieves impressive performance across classical IQA and IAA tasks, few-label IQA and other downstream tasks, showing promise as a foundational assessment model. 
\end{itemize}

\section{Related Work}
\label{sec:related_work}

\subsection{Image Quality Assessment} 

The rapid development of deep learning has sparked significant interest in their application for IQA.
Many researchers utilize CNN to solve the IQA problem with various effective techniques, including multi-level feature aggregation~\cite{li2018has}, adaptive quality prediction~\cite{su2020blindly}, and patch-to-picture learning~\cite{ying2020patches}, and unsupervised learning~\cite{saha2023re}.
Recently, transformer-based IQA methods~\cite{ke2021musiq,qin2023data,xu2024boosting,yu2024sf} show promising results in the IQA field, which can compensate for the non-local representation ability of CNN.
Despite these impressive breakthroughs, these methods often transfer models pre-trained on classification datasets, such as ImageNet~\cite{deng2009imagenet}, to IQA tasks, which may be suboptimal~\cite{li2023adaptive}. 
Q-Align~\cite{wu2023q} attempts to jointly perform IQA and IAA tasks, but it uses a huge language model and does not explicitly extract features of the two tasks through pre-training.
Our method can learn more effective representations through joint pre-training on quality-aesthetics image-text data, benefiting IQA tasks.


\subsection{Image Aesthetic Assessment} 

Image Aesthetic Assessment (IAA) aims to measure the aesthetic quality of images. 
With the advent of deep learning, IAA methods have evolved from hand-crafted feature extraction~\cite{datta2006studying,ke2006design,nishiyama2011aesthetic} to end-to-end feature learning, marking significant advancements in the IAA domain.
Various techniques have been developed to boost IAA task, including local and global feature integration~\cite{lu2015deep,hou2020object,shi2024improving,huang2024coarse,he2023eat}, graph network~\cite{she2021hierarchical,duan2022semantic}, knowledge distillation~\cite{hou2022distilling} 
and theme-aware learning~\cite{li2023theme,he2022rethinking}. 
Recently, there has been an emergence of multimodal IAA methods~\cite{zhang2020beyond,zhou2016joint,nie2023bmi,huang2024aesexpert} that incorporate text as auxiliary supervision. 
However, these methods necessitate the use of text during inference, limiting their flexible application since text is often not easily available. 
Our method overcomes this limitation by conducting vision-language pre-training firstly to learn effective representation. The pre-trained model can be flexibly applied to the IAA field using only images.

\subsection{Vision-Language Models}

Vision-Language Models (VLMs)~\cite{radford2021learning,yao2021filip,yu2022coca,sun2023eva} introduce the contrastive learning strategy to acquire image-text correspondences from large-scale image-text pairs.
VLMs have exhibited promising results across multiple tasks, including IQA~\cite{wang2023exploring,zhang2023blind} and IAA~\cite{hentschel2022clip,sheng2023aesclip}. 
Recently, the Multimodal Large Language Models (MLLMs) have garnered increasing research interest, exhibiting remarkable prowess in comprehending image content and reasoning through complex instructions~\cite{liu2024visual,zhu2023minigpt,li2023otter,ye2023mplug,bai2023qwen}. Most existing MLLMs achieve this by integrating image features with LLM tokens, subsequently fine-tuning the LLM via multimodal instruction tuning. During inference, MLLMs can reason with given images and user instructions, generating text responses by leveraging world knowledge learned during pre-training.

\section{UniQA: MLLMs-assisted Unified Pre-training}
\label{sec:uniqa}


In this section, we first present some preliminaries of related models. 
We then describe the process of constructing a unified image-text dataset about quality and aesthetics, with the assistance of MLLMs (Section~\ref{sec:captioning} and \ref{sec:data-refine}). 
We use this dataset for vision-language pre-training to construct UniQA.

\subsection{Preliminaries}
\label{sec:preliminaries}

Vision-language pre-training aims to achieve comprehensive cross-modality understanding by training on web-scale image-text datasets. Benefiting from this large-scale pre-training, CLIP, a prominent VLM, has demonstrated great promise to assist a broad scope of vision tasks.
Specifically, CLIP comprises an image encoder $f$ and a text encoder $g$, both jointly trained to establish a shared latent space for image and text through contrastive learning.

Given a batch of $N$ paired images and texts $\{x_I^i, x_T^i\}_{i=1}^N$, 
CLIP extracts image features $\bm{I} = \{f(x_I^i)\}_{i=1}^N$ and text features $\bm{T} = \{g(x_T^i)\}_{i=1}^N$ with corresponding encoders. 
During pre-training, CLIP seeks to maximize the cosine similarity of paired image and text features, while minimizing the similarity of unmatched pairs.
The contrastive learning objective can be formulated as:
\begin{equation}
\begin{aligned}
\mathcal{L}_\text{image} = - \mathbb{E}_{I_i \sim \bm{I}} \left[ \log \frac{\exp (I_i^\top T_i/\tau)} {\sum_{j=1}^{N} \exp (I_i^\top T_j/\tau)}\right] \\
\mathcal{L}_\text{text} = - \mathbb{E}_{T_i \sim \bm{T}} \left[\log \frac{\exp (T_i^\top I_i/\tau)} {\sum_{j=1}^{N} \exp (T_i^\top I_j/\tau)}\right]
\label{eq:contrastive_learning}
\end{aligned}
\end{equation}
where the $I_i$ and $T_i$ are the $i$-th features in the batch, and $\tau$ is the temperature parameter. The final contrastive learning loss can be obtained by taking the average: $\mathcal{L} = (\mathcal{L}_\text{image} + \mathcal{L}_\text{text})/2$. 
With this training strategy, CLIP can generate aligned features for paired image-text samples.

\subsection{Quality- and Aesthetics-related Captioning}
\label{sec:captioning}

\begin{figure}[t]
\centering
\includegraphics[scale=0.35]{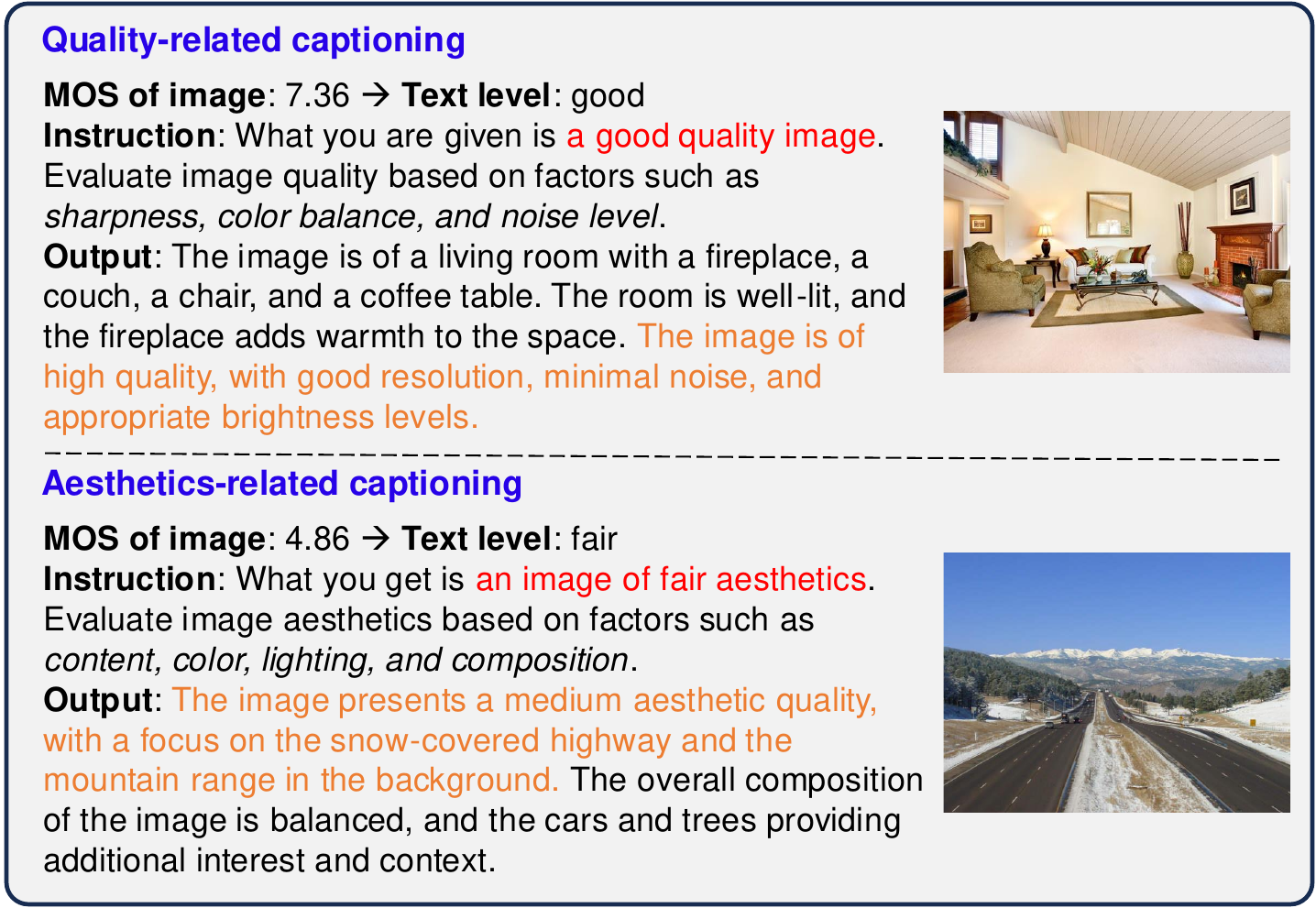}
\caption{
Generating quality- and aesthetics-related captions via MLLMs and our MOS-guided task-specific prompts. The \textcolor[rgb]{1,0,0}{red text} refers to MOS-based text guidance. The \textcolor[rgb]{1,0.5,0.25}{orange text} highlights the quality- and aesthetics-related text.
}
\label{fig:prompt}
\end{figure}

\begin{figure*}[t]
\centering
\includegraphics[scale=0.54]{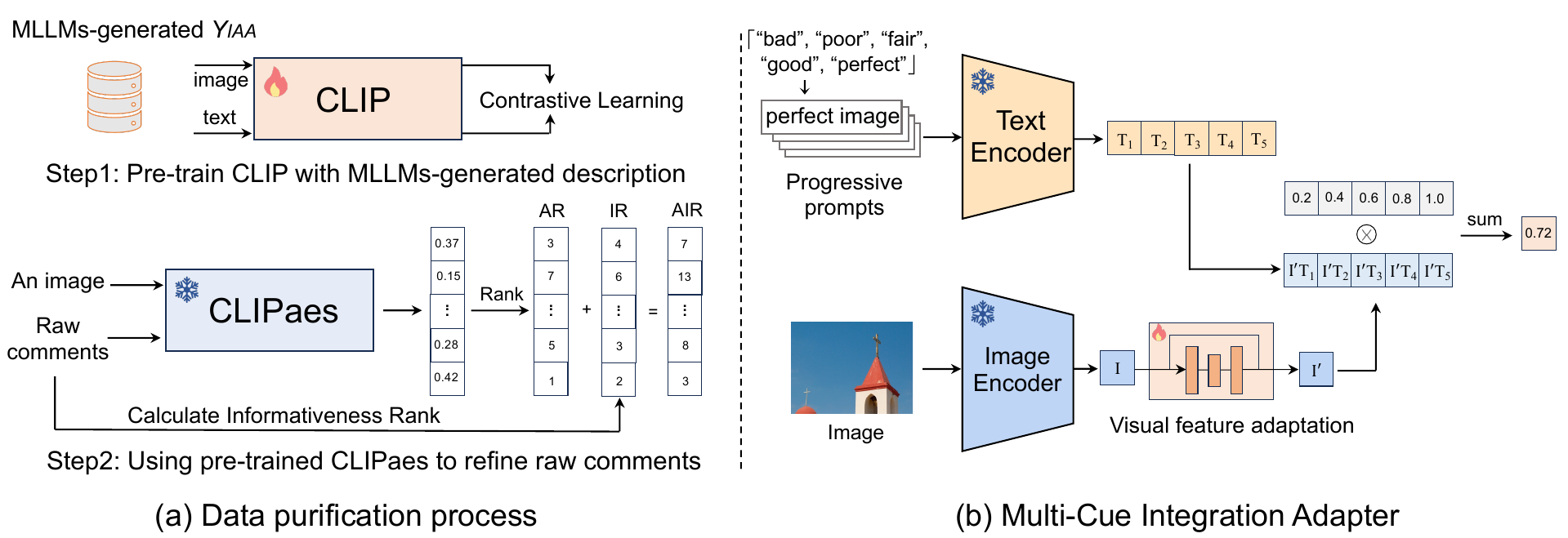}
\caption{(a) Data purification process: we  pre-train CLIP using generated aesthetic captions data $Y_{IAA}$ and then use the pre-trained $\mathrm{CLIP}_{aes}$ to purify data. (b) The proposed adapter: we employ progressive prompts, \{$\texttt{bad}$, $\texttt{poor}$, $\texttt{fair}$, $\texttt{good}$, $\texttt{perfect}$\} with ``image'', to prompt the frozen UniQA and a lightweight trainable module to adjust visual features.
}
\label{fig:model}
\end{figure*}

In order to achieve vision-language pre-training in the field of image assessment, we need to generate text for IQA and IAA datasets since IQA datasets lack text and IAA datasets contain noisy text. Recently, MLLMs have shown advanced performance, so we can use them to generate high-quality textual data for images. Previous studies~\cite{wu2023qbench,huang2024aesbench} have highlighted that it is challenging for MLLMs to directly and accurately perceive the quality and aesthetics of input images, often resulting in positively skewed expressions and strong hallucinations. 
Thus, to obtain correct and detailed descriptions about quality and aesthetics, as shown in Fig.~\ref{fig:prompt}, we design \textbf{MOS-guided task-specific prompts} to instruct MLLMs:
\begin{equation}
    Y_t \sim M_T(x_I, P_t|G).
\end{equation}
where $M_T$ denotes the used MLLM, $G$ is the MOS-based text guidance, $P_t$ is the task-specific prompt, $Y_t$ represents the generate caption.
To obtain $G$, we divide images into $5$ levels based on MOS, \textit{i.e.}, \{$\texttt{bad}$, $\texttt{poor}$, $\texttt{fair}$, $\texttt{good}$, $\texttt{perfect}$\}~\cite{clive,zhang2023blind}. If an image's MOS ranks in the top $20\%$ of the score range, its level is assigned to $\texttt{perfect}$. This approach harmonizes IQA and IAA datasets with different MOS scales, alleviating the MOS biases of different datasets~\cite{zhang2021uncertainty}.
Additionally, $P_t$ is customized for IQA ($P_{IQA}$) and IAA ($P_{IAA}$) tasks, respectively. 
As shown in Fig.~\ref{fig:prompt}, $P_{IQA}$ involves \textit{sharpness, color balance, and noise level}~\cite{chandler2013seven}, while $P_{IAA}$ includes \textit{content, color, lighting, and composition}~\cite{deng2017image}.
With these designs, $M_T$ is guided towards image assessment and we can obtain generated caption datasets $Y_{IQA}$ and $Y_{IAA}$ for IQA and IAA tasks, respectively. For simplicity and cost-effectiveness, we use open-source LLaVA~\cite{liu2023improved} as the captioner. We also experiment with the effects of different MLLMs on model performance (Table~\ref{tab:ablation}).

\subsection{Data Purification Strategy} 
\label{sec:data-refine}

In addition to the generated aesthetic captions $Y_{IAA}$, there are also IAA datasets with captions commented by humans~\cite{avacaption}, which directly reflect human aesthetic feelings.
Incorporating comments from various people can offer a more comprehensive description of image aesthetics. 
However, 
it may introduce noise to the data, as individuals may provide comments unrelated to image aesthetics. 
To address this issue, we propose a novel data purification strategy to refine raw captions in the original dataset. This process is illustrated in Fig.~\ref{fig:model}(a). 


Specifically, we introduce \textbf{{{Aesthetics-relevance and Informativeness Rank (AIR)}}} to measure the quality of text corresponding to an image. The AIR consists of {Aesthetics-relevance Rank (AR)} and {Informativeness Rank (IR)}. 
To obtain AR, we first pre-train a CLIP model with generated aesthetic data $Y_{IAA}$ to get an aesthetics-aware CLIP model, denoted as $\mathrm{CLIP_{aes}}$. 
Then, we employ it to measure the aesthetics relevance score ($s_{A}$) for an image-text pair. Given an image with $n$ captions, AR can be defined as:
\begin{equation}
\mathrm{AR} = \mathrm{Rank}(s_A^1 \cdots s_A^{n}), \quad  s_A^i=\mathrm{CLIP_{aes}}(x_I,x_T^i),
\label{eq:ar}
\end{equation}
where $s_A^i$ represents the aesthetics relevance score between the $i$-th caption $x_T^i$ and its corresponding image $x_I$. Note that $\mathrm{AR}$ consists of $\textit{long integers}$ that represent the rank of a caption after sorting by $s_A$.
For an image with $n$ textual captions, IR can be expressed as:
\begin{equation}
\mathrm{IR} =\mathrm{Rank}(s_I^1, \cdots, s_I^{n}),\quad s_I^i=\mathrm{InfoMeasure}(x_T^i), 
\label{eq:ir}
\end{equation}
where $\mathrm{InfoMeasure}(\cdot)$ can output the informativeness score ($s_I$) of an input sentence. The informativeness score of the text can be measured by various methods, such as sentence length or Shannon entropy~\cite{shannon1948mathematical}. Here we simply use the length of the sentence as the informativeness score and discuss other methods in ablation studies (Table~\ref{tab:ablation}). 
As a result, the AIR between an image and $n$ captions is:
\begin{equation}
\mathrm{AIR} =\mathrm{Rank}((\alpha \mathrm{AR}^{1} + \beta \mathrm{IR}^{1}), \cdots, (\alpha \mathrm{AR}^{n} + \beta \mathrm{IR}^{n})),
\label{eq:air}
\end{equation}
where $\alpha$ and $\beta$ are used to balance $\mathrm{AR}$ and  $\mathrm{IR}$ to purify the data more flexibly. In implementation, we set them to one for simplicity.
We select captions with Top-K ranking AIR to construct a high-quality 
aesthetic caption dataset, denoted as $Y_{
IAA}^{+}$.
This strategy ensures that the text is both aesthetically relevant and informative, thereby improving the quality and richness of the raw dataset.

\subsection{Unified Vision-Language Pre-training} 
\label{sec:pre-training}
So far, we have gotten a high-quality image-text dataset about quality and aesthetics, $Y=Y_{IQA} \cup Y_{IAA} \cup  Y_{IAA}^{+}$. Based on it, we pre-train CLIP using Equation~\ref{eq:contrastive_learning} to obtain our UniQA. In this way, the model learns general perceptions of image quality and aesthetics, 
which can provide potent assessment priors and thus can be effectively applied to both IQA and IAA tasks.

\section{Adapting UniQA for IQA and IAA}
\label{sec:adpater}


The pre-trained UniQA contains extensive perception information, which can facilitate downstream assessment tasks in a zero-shot or supervised manner. In this section, we further propose a meticulously designed adapter  and prompt ensemble strategy to enhance the model's performance.

\subsection{Multi-Cue Integration Adapter} 
\label{sec:adapter}

During pre-training, the model aligns image and assessment-related captions, empowering it with strong comprehension of image quality and aesthetics.
With this foundation model, we can slightly adjust the visual features, efficiently adapting it to score-based image assessment tasks.
To this end, we introduce a lightweight adapter, namely the \textbf{{{Multi-Cue Integration Adapter}}}, to adapt visual features and inject rich cues for fine-tuning downstream tasks. 
The adapter consists of two key processes: visual feature adaptation and multi-cue integration prediction.

\begin{table*}[t]
  \caption{
  Results on IQA datasets. \textbf{Black} and \second{blue} numbers in bold represent the best and second best, respectively. $^{\dag}$ denotes unfreezing the UniQA backbone for fine-tuning.
  Higher SRCC and PLCC imply better performance
  } 
  
    \centering
        \begin{tabular}{lcccccccc|cccccc}
      \toprule[1.5pt]
      
   & \multicolumn{2}{c}{LIVE} & \multicolumn{2}{c}{TID2013} & \multicolumn{2}{c}{CSIQ}  & \multicolumn{2}{c|}{KADID} & \multicolumn{2}{c}{CLIVE} & \multicolumn{2}{c}{KonIQ}  & \multicolumn{2}{c}{SPAQ}\\
   \cmidrule{2-15}   
   Method & 
    \multicolumn{1}{c}{SRCC} & \multicolumn{1}{c}{PLCC}  & \multicolumn{1}{c}{SRCC} & \multicolumn{1}{c}{PLCC}  & \multicolumn{1}{c}{SRCC} & \multicolumn{1}{c}{PLCC}& \multicolumn{1}{c}{SRCC} & \multicolumn{1}{c|}{PLCC}& \multicolumn{1}{c}{SRCC} & \multicolumn{1}{c}{PLCC}& \multicolumn{1}{c}{SRCC} & \multicolumn{1}{c}{PLCC}& \multicolumn{1}{c}{SRCC} & \multicolumn{1}{c}{PLCC}\\

\midrule

BRISQUE~\cite{mittal2012no} & 0.436 & 0.459 & {0.626}  & {0.571} & {0.812} & {0.748} & {0.528} & {0.567} & {0.629} & {0.629} & 0.681 & 0.685 & 0.809 & 0.817 \\



WaDIQaM~\cite{bosse2017deep} & {0.960}  & {0.955} & {0.835}  & {0.855} & {0.852} & {0.844} & {0.739} & {0.752} & {0.682} & {0.671} & 0.804 & 0.807 & 0.840 & 0.845 \\
 
DBCNN~\cite{zhang2018blind} & {0.968}  & {0.971} & {0.816}  & {0.865} & {0.946} & {0.959} & {0.851} & {0.856} & {0.851} & {0.869} & 0.875 & 0.884 & 0.911 & 0.915 \\

MetaIQA~\cite{zhu2020metaiqa} & {0.960}  & {0.959} & {0.856}  & {0.868} & {0.899} & {0.908} & {0.762} & {0.775} & {0.802} & {0.835} & 0.850 & 0.887 & - & - \\

PaQ-2-PiQ~\cite{ying2020patches} & {0.959}  & {0.958}  & {0.862}  & {0.856} & {0.899} & {0.902} & {0.840} & {0.849} & {0.844} & {0.842} & 0.872 & 0.885 & - & - \\

HyperIQA~\cite{su2020blindly} & {0.962}  & {0.966} & {0.840}  & {0.858} & {0.923} & {0.942} & {0.852} & {0.845} & {0.859} & {0.882} & 0.906 & 0.917 & 0.911 & 0.915 \\

TReS~\cite{golestaneh2022no} & {0.969}  & {0.968} & {0.863}  & {0.883} & {0.922} & {0.942} & {0.859} & {0.858} & {0.846} & {0.877} & 0.915 & 0.928 & {-} & - \\

MUSIQ~\cite{ke2021musiq} & {0.940}  & {0.911} & {0.773}  & {0.815} & {0.871} & {0.893} & {0.875} & {0.872} & {0.702} & {0.746} & 0.916 & 0.928 & {0.918} & 0.921 \\

    DACNN~\cite{pan2022dacnn} & 0.978 & 0.980 & {0.871}  & {0.889} & {0.943} & {0.957} & {0.905} & {0.905} & {0.866} & {0.884} & 0.901 & 0.912 & {0.915} & 0.921 \\
    
     DEIQT~\cite{qin2023data} & {0.980} &  {0.982} & {0.892} &  {0.908} & {0.946} & {0.963} & 0.889 & 0.887  & {0.875} & {0.894} & {0.921} & {0.934} & {0.919} & {0.923} \\

     LIQE~\cite{zhang2023blind} & 0.970 & 0.951  & - & & 0.936 & 0.939 & {0.930} & {0.931} & \second{0.904} & {0.911} & 0.919 & 0.908 & - & - \\ 
     
     Re-IQA~\cite{saha2023re} & 0.970 & 0.971 & 0.804 &  0.861 & {0.947} & {0.960} &{0.872} & 0.885 & 0.840 &{0.854} & {0.914} &0.923 &0.918 & 0.925  \\ 

     LoDA~\cite{xu2024boosting} & 0.975 & 0.979 & 0.869 &  0.901 & - & - & {0.931} & {0.936} & 0.876 &{0.899} & {0.932} & \second{0.944} & \textbf{0.925} & \second{0.928} \\

     Q-Align~\cite{wu2023q} & 0.913 &  0.919 & - &  - & 0.915 & 0.936 & {0.869} & {0.927} & \textbf{0.931} & \textbf{0.921} & \second{0.935} & {0.934} & - & - \\
     
    \midrule
    \rowcolor{mygray} Ours & \second{0.981} & \second{0.983} & \second{0.916} & \second{0.931} & \second{0.963} & \second{0.973} & \second{0.940} & \second{0.943} & {0.890} & {0.905} & {0.933} & {0.941} & \second{0.924} & \second{0.928} \\

    \rowcolor{mygray} Ours$^{\dag}$ & \textbf{0.985} & \textbf{0.985}  & \textbf{0.950} & \textbf{0.959} & \textbf{0.977} & \textbf{0.980} & \textbf{0.966} & \textbf{0.968} & {0.902} & \second{0.914} & \textbf{0.940} & \textbf{0.948} & \textbf{0.925} & \textbf{0.929} \\

    \bottomrule
    \end{tabular}
  \label{tab:iqa-sota}
\end{table*}

\noindent \textbf{Visual Feature Adaptation.} 
We add a learnable residual module following the image encoder to adjust the visual features so as to adapt to specific assessment datasets. We optimize this module while keeping the image and text backbones frozen, enabling parameter-efficient tuning. The adapter is illustrated in Fig.~\ref{fig:model}(b). Let $I$ denote the image features extracted from the frozen image encoder, the visual feature adaptation process can be expressed as:
\begin{equation}
\begin{aligned}
I^{\prime}=\mathrm{Normalize}(\mathrm{Adapter}(I)+I)
\label{eq:scale-fusion}
\end{aligned}
\end{equation}
where the $\mathrm{Adapter(\cdot)}$ consists of two fully connected layers with a $\mathrm{ReLU}$ activation in between, and $I^{\prime}$ represents the adapted visual features.

\noindent \textbf{Multi-cue Integration Prediction.} A straightforward approach to incorporating CLIP into image assessment is to use the ``$\texttt{good image}$'' as an anchor and take the cosine similarity between the text anchor and a given image as the assessment score. However, this method shows two shortcomings: (1) using the absolute value of similarity as score may not be optimal because it only reflects the semantic similarity between images and texts~\cite{wu2023qbench,wang2023exploring}; (2) a single prompt may not fully leverage the extensive knowledge of the pre-trained model. Thus, we propose to utilize versatile cues to comprehensively explore the power of the pre-trained UniQA and convert absolute similarity scores into relative values for weighting.

Specifically, we utilize the prompt template ``\{level\} image'' and  five text levels ($\texttt{bad}$, $\texttt{poor}$, $\texttt{fair}$, $\texttt{good}$, $\texttt{perfect}$), \textit{i.e.}, ``\textit{Multi-cue}'', to construct prompts. 
Next, we calculate the cosine similarity between the normalized text features $\{T_i\}_{i=1}^{5}$ of five prompts and adapted visual features $I^{\prime}$, and then use the $\mathrm{Softmax(\cdot)}$ to obtain the related value of five image-text correspondence. These related values will weight the predefined score levels to get the final assessment score. 
The process is described in Fig.~\ref{fig:model}(b) and can be formulated as follows:
\begin{equation}
q =\sum_{i=1}^5\frac{c_i \exp({{I^{\prime}}^{\top} T_i}/\tau)}{\sum_{j=1}^5 \exp({{I^{\prime}}^{\top} T_i}/\tau)},
\label{eq:Attention}
\end{equation}
where $\{c_i\}_{i=1}^{5}$ are scores of text levels with progressive values that are set to $\{0.2, 0.4, 0.6, 0.8, 1.0\}$. Note that the scores are learnable parameters and can be dynamically adjusted based on the datasets. $\tau$ is the temperature parameter and $q$ is the assessment score of the given image.

\begin{table}[!t]
    \footnotesize
    \caption{The assessment-oriented ensemble prompts used in zero-shot and few-label learning}
    \resizebox{0.49\textwidth}{!}{
    \begin{tabular}{lccc}
    \toprule
    Task & Prompt \\
    \midrule
    \multirow{8}{*}{\makecell{CLIVE\\KonIQ\\LIVE}}
     & \{$\texttt{bad}$, $\texttt{poor}$, $\texttt{fair}$, $\texttt{good}$, $\texttt{perfect}$\} with $\texttt{image}$ \\ \cmidrule{2-2}
     
     & \{$\texttt{extremely blurry}$, $\texttt{blurry}$, \\
     &$\texttt{fair}$, $\texttt{sharp}$, $\texttt{extremely sharp}$\} with $\texttt{image}$ \\ \cmidrule{2-2}
     
     &  \{$\texttt{extremely noisy}$, $\texttt{noisy}$, $\texttt{fair}$,\\
     & $\texttt{noise-free}$, $\texttt{extremely noise-free}$\} with $\texttt{image}$ \\ \cmidrule{2-2}
     
     & \{$\texttt{extremely low-quality}$, $\texttt{low-quality}$, $\texttt{fair}$, \\
     & $\texttt{high-quality}$, $\texttt{extremely high-quality}$\} with $\texttt{image}$ \\ \midrule
     \multirow{2}{*}{AGIQA-3K}
     & \{$\texttt{bad}$, $\texttt{poor}$, $\texttt{fair}$, $\texttt{good}$, $\texttt{perfect}$\} with $\texttt{image}$ \\
     & \{$\texttt{bad}$, $\texttt{poor}$, $\texttt{fair}$, $\texttt{good}$, $\texttt{perfect}$\} with $\texttt{content}$ \\
    \bottomrule
    \end{tabular}
    \label{tab:prompt_es}
    }
\end{table}

\subsection{Assessment-oriented Prompt Ensemble} 
\label{sec:prompt}

We introduce an \textbf{assessment-oriented prompt ensemble} strategy, which incorporates more image assessment-related prompt groups to derive the final assessment score, thereby achieving a more comprehensive understanding of image quality and aesthetics.
For instance, we can use \textit{e.g.}, \{$\texttt{extremely blurry}$, $\texttt{blurry}$, $\texttt{fair}$, $\texttt{sharp}$, $\texttt{extremely sharp}$\} as another five text levels. Now, the final assessment score $q_f$ is the average of all prompt groups and it can be described as:
\begin{equation}
\begin{aligned}
q_f &=\frac{\sum_{i=1}^m q_i}{m},
\label{eq:QCIM}
\end{aligned}
\end{equation}
where $m$ denotes the number of prompt groups.
This strategy can more fully utilize the multi-modal understanding capabilities of UniQA and demonstrates non-negligible performance improvements in zero-shot (Table~\ref{tab:cross-dataset}) and few-label learning (Table~\ref{tab:few-shot-sota}). The details of ensemble prompts are represented in Table~\ref{tab:prompt_es}. Note that the prompts for AGIQA-3K differ from other IQA datasets, as distortions in AI-generated images is different from those in authentic images. For instance, authentic image distortions may stem from camera shake, whereas AI-generated image distortions typically involve meaningless content and distorted poses. Thus, ``content'' is used to prompt UniQA for the AGIQA-3K dataset.

\begin{table*}[t]
\centering
    \begin{minipage}[!t]
    {0.23\textwidth}
    \caption{Results on AVA dataset}
    \label{table:5}
    \centering
    \begin{tabular}{lcccc}
    \toprule
 Method &SRCC &PLCC \\
    
    \midrule
    NIMA~\cite{talebi2018nima} & 0.612 &0.636 \\
    MaxViT~\cite{tu2022maxvit} &0.708 &0.745 \\
    MUSIQ~\cite{ke2021musiq}   & 0.726 &0.738 \\
    MLSP~\cite{hosu2019effective}    & 0.756 & {0.757} \\
    TANet~\cite{he2022rethinking}    & 0.758 & {0.765} \\
     MILNet~\cite{shi2024improving}    &  {0.732} &  {0.753} \\
    EAT~\cite{he2023eat} & 0.759 & {0.770} \\
    VILA~\cite{ke2023vila} & {0.774} & {0.774} \\
    \midrule
       \rowcolor{mygray} Ours    &\second{0.776} & \second{0.776} \\
       \rowcolor{mygray} Ours$^{\dag}$    &\textbf{0.782} & \textbf{0.782} \\
    \bottomrule
    \end{tabular}
    \label{tab:ava-sota}
    \end{minipage}
    \hspace{0.05\textwidth}
    \begin{minipage}[!t]{0.24\textwidth}
        \caption{Results on AADB dataset}
        \begin{tabular}{lcc}
        \toprule
        Method &SRCC &PLCC \\
    \midrule
        NIMA~\cite{talebi2018nima} & 0.708 &0.711 \\
        MLSP~\cite{hosu2019effective}    & 0.725 &0.726 \\
        MUSIQ~\cite{ke2021musiq}    & 0.706 & {0.712} \\
        PA-IAA~\cite{li2020personality}    & 0.720 & {0.728} \\
        HIAA~\cite{li2022psychology}    & 0.739 & - \\
        TANet~\cite{he2022rethinking}    & 0.738 & {0.737} \\
        Celona  \textit{et al.}~\cite{celona2022composition}    & 0.757 & {0.762} \\
        TAVAR~\cite{li2023theme}     &{0.761} &{0.763} \\
    \midrule
      \rowcolor{mygray}  Ours    & \second{0.786} & \second{0.787} \\
      \rowcolor{mygray}  Ours$^{\dag}$    & \textbf{0.788} & \textbf{0.791} \\
    \bottomrule
    \end{tabular}
    \label{tab:aadb-sota}
    \end{minipage}
    \hspace{0.08\textwidth}
    \begin{minipage}[!t]{0.22\textwidth}
        \caption{
        Results on BAID dataset
        }
    		\begin{tabular}{lcc}
			\toprule
    Method &SRCC &PLCC \\ \midrule
    NIMA~\cite{talebi2018nima} & 0.393 &0.382 \\
    MP$_{ada}$~\cite{sheng2018attention} &0.437 &0.425 \\
    MLSP~\cite{hosu2019effective}   & 0.441 &0.430 \\
    BIAA~\cite{zhu2020personalized}    & 0.389 & {0.376} \\
    TANet~\cite{he2022rethinking}    & 0.453 & {0.437} \\
     SAAN~\cite{yi2023towards}    &  {0.473} &  {0.467} \\
    TSC-Net~\cite{wang2023tsc} & 0.480 & {0.479} \\
    EAT~\cite{he2023eat} & \second{0.486} & {0.495} \\
    \midrule
       \rowcolor{mygray} Ours    &\textbf{0.487} & \textbf{0.528} \\
       \rowcolor{mygray} Ours$^{\dag}$    &{0.484} & \second{0.502} \\
            \bottomrule
		\end{tabular}
    \label{tab:baid-sota}
    \end{minipage}
\end{table*}


\section{Experiments}
\label{sec:exp}

\subsection{Datasets}

We employ the IQA dataset FLIVE~\cite{ying2020patches} and the IAA dataset AVA~\cite{ava} for quality- and aesthetics-related captioning, and AVA-Captions~\cite{avacaption} to provide authentic aesthetic comments. We evaluate the performance on typical IQA and IAA datasets, including seven IQA datasets and three IAA datasets. We also evaluate the generalization ability of UniQA on three downstream IQA datasets.

\textbf{IQA Dataset}. For the IQA task, four synthetic datasets, including LIVE~\cite{live}, CSIQ~\cite{csiq}, TID2013~\cite{tid2013}, KADID~\cite{kadid}, and three authentic datasets of CLIVE~\cite{clive}, KonIQ~\cite{koniq}, SPAQ~\cite{spaq}, are used for performance evaluation. FLIVE~\cite{ying2020patches} is an authentic IQA dataset that contains 39,810 images.
We employ three datasets to evaluate the generalization capability of our UniQA, including an AI-generated IQA dataset AGIQA-3K~\cite{li2023agiqa3k}, an underwater IQA dataset UWIQA~\cite{uwiqa} and an enhanced colonoscopy image quality assessment dataset ECIQAD~\cite{yue2023perceptual}. The details of the used datasets are shown in Table~\ref{tab:iqa_iaa_data}.

\textbf{IAA Dataset}. For the IAA task, we conduct experiments on three datasets, including AVA~\cite{ava}, AADB~\cite{aadb} and BAID~\cite{yi2023towards} datasets. AVA comprises 250k images, with the test set of 19,928 images. 
AADB dataset consists of 10,000 images in total, with 8,500 images for training, 500 images for validation, and 1,000 images for testing. BAID is a large-Scale artistic image aesthetics Assessment dataset, with 60,337 images in tatol. We follow the standard data split and use 53,937 images for training and 6,400 images for testing.


\begin{table}[!t]
    \centering
    \caption{Details of different IQA and IAA datasets. Dist.~No. indicates the number of distortion types}
    \begin{tabular}{lcccc}
    \toprule
    Dataset & Task &  Type  &  Size & Dist. No. \\
    \midrule
    LIVE~\cite{live} & IQA & Synthetic & 799 & 5 \\
    CSIQ~\cite{csiq} & IQA & Synthetic & 866 & 5 \\
    TID2013~\cite{tid2013} & IQA & Synthetic & 3,000 & 24 \\
    KADID~\cite{kadid} &IQA & Synthetic & 10,125 & 25 \\ 
    CLIVE~\cite{clive} & IQA & Authentic & 1,162 & - \\
    KonIQ~\cite{koniq} & IQA & Authentic & 10,073 & - \\
    SPAQ~\cite{spaq} & IQA & Authentic & 11,000 & - \\
    FLIVE~\cite{ying2020patches} & IQA & Authentic & 39,810 & - \\
    AGIQA-3K~\cite{li2023agiqa3k} & IQA & Authentic & 2,982 & - \\
    UWIQA~\cite{uwiqa} & IQA & Authentic & 890  & - \\
    ECIQA~\cite{yue2023perceptual} & IQA & Authentic & 2400 & -\\
    \midrule
    AVA~\cite{ava} & IAA & Authentic & 250,000 & - \\
    AADB~\cite{aadb} & IAA & Authentic & 10,000 & -\\
    BAID~\cite{yi2023towards} & IAA & Authentic & 60,337 & -\\
    \bottomrule
    \end{tabular}
    \label{tab:iqa_iaa_data}
\end{table}

\textbf{AVA-Captions}. AVA-Captions offer multiple human-annotated comments for each AVA image. To avoid potential data leakage, we  strictly follow the official data split of AVA, results in a pre-training image-text dataset comprising 234,090 images paired with 3.0 million captions.

\subsection{Evaluation Criteria}

We employ Spearman's Rank-order Correlation Coefficient (SRCC) and Pearson’s Linear Correlation Coefficient (PLCC) as criteria to measure the performance of IQA and IAA models. They reflect the prediction monotonicity and prediction accuracy of the model, respectively. Both SRCC and PLCC range from 0 to 1. Higher values of SRCC and PLCC indicate better performance. For each IQA dataset, 80\% of the images are used for training and the remaining 20\% for testing.
We repeat this process 10 times to mitigate the performance bias and the medians of SRCC and PLCC are reported. For the IAA datasets, we follow the standard data splits.

\begin{table}[!t]
\centering
\caption{SRCC on the cross datasets validation. $^{*}$ denotes using ensemble prompts. The results of other methods are pre-trained on FLIVE}
    {
        \begin{tabular}{lcc|c}
        \toprule
{Method}  & CLIVE & KonIQ & AGIQA-3K  \\
        \midrule
        DBCNN~\cite{zhang2018blind}   & 0.724 & 0.716 & 0.645 \\
        PaQ-2-PiQ~\cite{ying2020patches}   & 0.738 & 0.755 & 0.502  \\
        HyperIQA~\cite{su2020blindly}  & {0.735} & 0.758 & 0.629  \\
        TReS~\cite{golestaneh2022no}  &  0.740 & {0.713} & 0.646 \\
        DEIQT~\cite{qin2023data}  &  \second{0.781} & \second{0.733} & -  \\
          CLIP$^{*}$~\cite{radford2021learning}  & {0.746} & {0.592} & {0.646} \\
        \midrule
        \rowcolor{mygray}  Ours  & {0.638} & {0.667} & \second{0.744} \\
    \rowcolor{mygray} 	Ours$^{*}$  & \textbf{0.790} & \textbf{0.806} & \textbf{0.752} \\
        \bottomrule
    \end{tabular}}
\label{tab:cross-dataset}
\end{table}

\subsection{Implementation Details}
\label{sec:imp_detail}

In this section, we introduce the details of large-scale assessment-related text data generation and purification, multimodal pre-training to build UniQA, and fine-tuning  UniQA for IQA and IAA tasks.

\textbf{Text data generation and purification.} We use LLaVA-1.5-7B~\cite{liu2024visual,liu2023improved} as the multimodal large language model (MLLM) for captioning. We generate three captions for each IQA image and one caption for each IAA image, resulting 119,421 generated IQA captions and 234,090 IAA captions. We use these high-quality generated IAA data as metadata to purify the raw AVA-Captions dataset. We set $K=4$ to refine the AVA-Captions dataset. In the end, we obtain 273,897 images, and paired with 1,240,915 comments. Each image has at least three captions.

\textbf{Multimodal pre-training.} We adopt CLIP-B/16~\cite{radford2021learning} as the vision-language model (VLM) for pre-training. We follows the same pre-training strategy as CLIP to train our UniQA.
We use the Adam optimizer~\cite{kingma2014adam} with a learning rate of 5e-6 and weight decay of 0.2. The training is performed for 5 epochs with a batch size of 960, and it is resource-efficient, taking \textit{less than an hour per run} on four A100 GPUs.

\textbf{Fine-tuning UniQA for downstream assessment.} 
We use Adam optimizer and MSE loss to fine-tune the pre-trained UniQA.
We employ a learning rate of 2e-4 for the adapter and 2e-6 for the UniQA backbone if the backbone is not frozen. We follow the typical training strategy to fine-tune each dataset, including random cropping and random horizontal flipping. Since different datasets have different MOS scales, we scale their range to [0, 1]
through normalization. Considering the data scale, we train 50 epochs for LIVE, CSIQ and CLIVE datasets, 10 epochs for AVA and BIAD datasets, and 20 epochs for other datasets. During inference, we typically crop an input image into 10 image patches and take their average as the quality score of this image~\cite{su2020blindly,qin2023data}. We use the resolution of 224 × 224 for training and testing.
All fine-tuning experiments are performed on an A100 GPU.

\begin{table*}[!ht]
	\centering
	\caption{Applying UniQA to AIGC-3K~\cite{li2023agiqa3k}, UWIQA~\cite{uwiqa} and ECIQAD~\cite{yue2023perceptual} datasets. Our UniQA can be used as a foundational model for other downstream tasks and achieves SOTA  performance}
 \label{tab:generalization}
    \begin{subtable}{0.26\textwidth}
    \center
    \caption{Results on the AIGC-3K dataset}\label{tab:aigc3k}
    \begin{tabular}{l c c c}
    \toprule
    Method &SRCC &PLCC \\
        \midrule
            NIQE~\cite{mittal2012no} & 0.524 &0.567 \\
            DBCNN~\cite{zhang2018blind} & 0.821 & 0.876 \\
            HyperNet~\cite{su2020blindly} & 0.836 & 0.890 \\
            CLIPIQA~\cite{wang2023exploring}    & 0.843 & {0.805} \\
            Q-Align~\cite{wu2023q}     & 0.673 &0.691 \\
            MUSIQ~\cite{ke2021musiq}     & 0.826 & 0.866 \\
            PSCR~\cite{yuan2023pscr}     & 0.850 & 0.906 \\
            
        \midrule
          \rowcolor{mygray}  Ours    & \second{0.876} & \second{0.917} \\
          \rowcolor{mygray}  Ours$^{\dag}$    & \textbf{0.888} & \textbf{0.922} \\
        \bottomrule
    \end{tabular}
    \end{subtable}
    \hspace{0.03\textwidth}
    \begin{subtable}{0.25\linewidth}
    \center
    \caption{Results on the UWIQA dataset}\label{tab:uwiqa}
    \begin{tabular}{l c c c}
    \toprule
    Method &SRCC &PLCC \\
    \midrule
    FDUM~\cite{uwiqa}     & 0.694 & 0.689 \\
    UCIQE~\cite{yang2015underwater}     & 0.627 &0.626 \\
    URanker~\cite{guo2023underwater}     & 0.674 &0.663 \\ 
    UIQM~\cite{hou2024no}    & 0.595 & {0.589} \\
    UIQI~\cite{liu2023uiqi}     & 0.742 &0.741 \\
    CSN~\cite{liu2023uiqi}     & 0.784 &0.753 \\
    HPUIQA~\cite{liu2023uiqi}     & 0.796 &0.790 \\
    \midrule
    \rowcolor{mygray}  Ours    & \second{0.847} & \second{0.859} \\
    \rowcolor{mygray}  Ours$^{\dag}$     & \textbf{0.876} & \textbf{0.889} \\
    \bottomrule
    \end{tabular}
    \end{subtable}
    \hspace{0.035\textwidth}
    \begin{subtable}{0.26\linewidth}
    \center
    \caption{Results on the ECIQAD dataset}\label{tab:eciqad}
    \begin{tabular}{l c c c}
    \toprule
    Method &SRCC &PLCC \\
        \midrule
            BRISQUE~\cite{mittal2012no} & 0.436 & 0.459 \\
        BIQME~\cite{gu2017learning}     & 0.770 & 0.768 \\
        BPRI~\cite{min2017blind}     & 0.152 &0.181 \\
        FRIQUEE~\cite{ghadiyaram2017perceptual}     & 0.663 &0.656 \\ CIQA~\cite{chen2021perceptual}     & 0.738 &0.735 \\
        ECIQ~\cite{ke2021musiq}    & 0.839 & {0.842} \\
        LIQE~\cite{ke2021musiq}    & \second{0.878} & {-} \\
            
        \midrule
          \rowcolor{mygray}  Ours    & {0.873} & \second{0.887} \\
          \rowcolor{mygray}  Ours$^{\dag}$     & \textbf{0.918} & \textbf{0.928} \\
        \bottomrule
    \end{tabular}
    \end{subtable}
\end{table*}

\subsection{Main Results}

\noindent \textbf{Results on IQA task.} Table~\ref{tab:iqa-sota} reports the performance of the SOTA IQA methods on seven typical IQA datasets.  
Our method demonstrates impressive performance across a diverse range of datasets, fully confirming the effectiveness of our method in precisely characterizing image quality. We can also observe that unfreezing the UniQA backbone for training improves performance effect, demonstrating the powerful quality characterization capability of UniQA. And fine-tuning the adapter which uses only 0.2\% of the number of parameters required for unfreezing UniQA backbone, yields highly competitive performance compared to other methods.
This shows the efficiency and effectiveness of our proposed adapter.
Furthermore, our method demonstrates more substantial advancements on synthetic data compared to authentic data, particularly evident on the TID2013~\cite{tid2013} and KADID~\cite{kadid} with its diverse distortion types and large data sizes. We attribute this phenomenon to the enhanced commonsense knowledge and perception of visual quality acquired through extensive quality and aesthetics pre-training.

\noindent \textbf{Results on IAA task.} We report the experimental results on the AVA, AADB and BAID datasets in Table~\ref{tab:ava-sota}, Table~\ref{tab:aadb-sota} and Table~\ref{tab:baid-sota}, respectively. Given that the pre-trained model acquired a unified and robust image assessment perception, it can also achieve SOTA results after fine-tuning on these three datasets. 
These results validate that our method can be effectively applied to both IQA and IAA domains.

\begin{table*}[t]
\centering
\caption{SRCC results under data-efficient learning setting. $^{*}$ denotes using ensemble prompts}
{
\begin{tabular}{lccc|ccc|ccc}
\toprule
  & \multicolumn{3}{c|}{CLIVE} & \multicolumn{3}{c|}{KonIQ} & \multicolumn{3}{c}{LIVE} \\ 
\cmidrule{2-10}
 Method & 50      & 100     & 200    & 50      & 100     & 200    & 50     & 100    & 200    \\ 
\midrule
HyperIQA~\cite{su2020blindly}   & 0.648 & 0.725 & 0.790 & 0.615 & 0.710 & 0.776 & 0.892 & 0.912 & 0.929  \\ 
TReS~\cite{golestaneh2022no}      & 0.670 & 0.751 & 0.799 & 0.713 & 0.719 & 0.791 & 0.901 & 0.927 & {0.957}     \\
CLIP~\cite{radford2021learning}      & 0.664 & 0.721 & 0.733 & 0.736 & 0.770 & 0.782 & 0.896 & 0.923 & 0.941    \\ 
CLIPIQA~\cite{wang2023exploring}      & 0.646 & 0.611 & 0.642 & 0.579 & 0.620 & 0.667 & 0.633 & {0.724} & 0.784  \\
Re-IQA~\cite{saha2023re}      & 0.591 & 0.621 & 0.701 & 0.685 & 0.723 & 0.754 & 0.884 & 0.894 & 0.929    \\ 
DEIQT~\cite{qin2023data}      & 0.667 & 0.718 & 0.812 & 0.638 & 0.682 & 0.754 & 0.920 & {0.942} & 0.955  \\
GRepQ~\cite{srinath2024learning}      & {0.760} & {0.791} & {0.822} & \second{0.812} & {0.832} & {0.855} &  {0.926} & 0.937 & 0.953    \\ 
\midrule
\rowcolor{mygray} Ours    & \second{0.813}     & \second{0.836}     & \second{0.850}    & {0.772}     & \second{0.842}     & \second{0.870}     & \second{0.962}    & \second{0.956}    & \second{0.974}    \\ 
\rowcolor{mygray} Ours$^{*}$      & \textbf{0.828}     & \textbf{0.849}     & \textbf{0.853}    & \textbf{0.844}     & \textbf{0.860}     & \textbf{0.876}    & \textbf{0.963}    & \textbf{0.958}    & \textbf{0.976}    \\ 
\bottomrule
\end{tabular}
}
\label{tab:few-shot-sota}
\end{table*}

\subsection{Generalization Capability Validation}


\noindent \textbf{Cross dataset validation.} Table~\ref{tab:cross-dataset} evaluate the generalization capability of our model. 
We directly utilize the pre-trained UniQA and textual prompts for quality assessment. It is more challenging than other methods as the model isn't optimized on MOS labels.
As observed, our method achieves the best performance on these three datasets. 
Notably, our method shows excellent performance on AIGC images (AGIQA-3K~\cite{li2023agiqa3k}), which are markedly different from natural images. These results demonstrate the strong generalization capability of our UniQA.
Additionally, the UniQA outperforms the original CLIP significantly, proving the effectiveness of our unified pre-training.

\begin{figure*}[!t]
\centering
\includegraphics[scale=0.52]{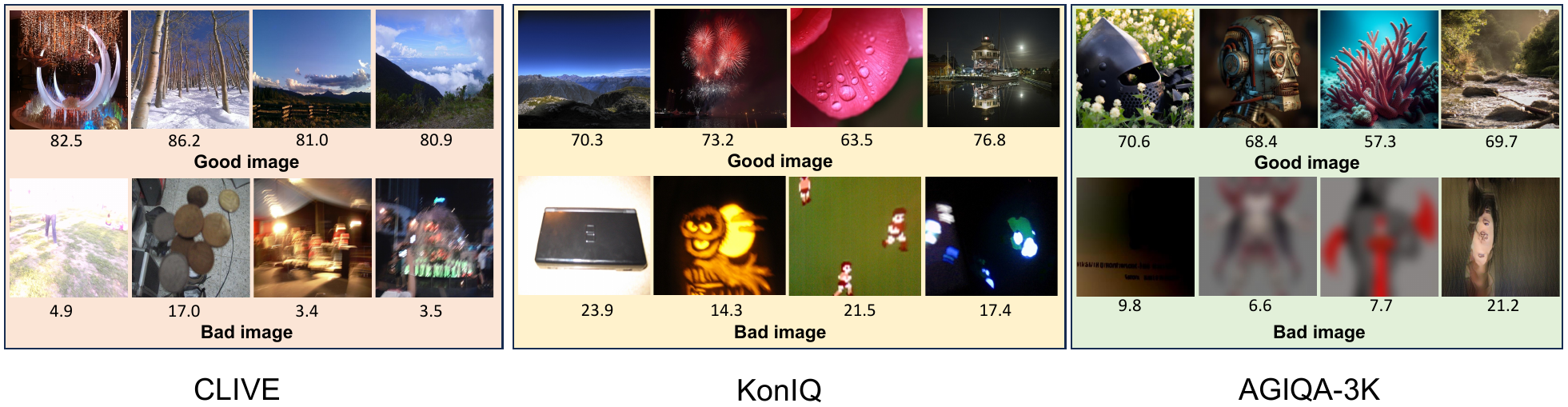}
\caption{
The image retrieval results on three dataset with varied prompts. The number below the image is its MOS label.
}
\label{fig:retri}
\end{figure*}

\begin{table}[t]
    \centering
    \caption{
    Ablation on IQA (CLIVE and KonIQ) and IAA (AVA) datasets with SRCC metrics
    }
 { 
   \begin{tabular}{ccc|ccc}
    \toprule 
     \multicolumn{3}{c|}{Ablation type}  & {CLIVE} & {KonIQ}& {AVA} \\
    \midrule
\multicolumn{6}{c}{{Ablation on different pre-training data}} \\
\midrule
\rowcolor{light-green} $Y_{IQA}$ & $Y_{IAA}$ & $Y_{IAA}^{+}$  & & &  \\
\rowcolor{light-green}  \mynocheck & \mynocheck & \mynocheck   & 0.865 &  0.907 	& 0.748     \\
\rowcolor{light-green} $\checkmark$  & \mynocheck & \mynocheck     & 0.871  & 0.914   & 0.755     \\
\rowcolor{light-green}  \mynocheck  & \checkmark &   \mynocheck  & 0.871  & 0.917  & 0.755     \\
\rowcolor{light-green}  \checkmark & \checkmark & \mynocheck &  0.874  & 0.918  & 0.756     \\
\rowcolor{light-green}  \mynocheck   & \mynocheck & \checkmark   & 0.875  & 0.928   & 0.773      \\
\rowcolor{light-green} \mynocheck  & \checkmark &  \checkmark    & 0.877  & 0.930   & 0.774     \\
\rowcolor{mygray} \checkmark  & \checkmark & \checkmark    & \textbf{0.890} & \textbf{0.933}  & \textbf{0.776}      \\
\midrule
\multicolumn{6}{c}{{Ablation on data purification strategy}} \\
\midrule
\rowcolor{light-yellow} \multicolumn{3}{c|} {w/o Strategy}        & 0.876 & 0.929   & 0.772     \\
\rowcolor{light-yellow} \multicolumn{3}{c|}{IR Strategy}        & 0.879  & 0.931   & 0.774     \\
\rowcolor{light-yellow} \multicolumn{3}{c|}{AR Strategy}        & 0.885 & 0.930   & 0.774     \\
\rowcolor{mygray} \multicolumn{3}{c|}{AIR Strategy}      & \textbf{0.890}  & \textbf{0.933} & \textbf{0.776}     \\
\midrule
\multicolumn{6}{c}{{Ablation on different IR strategies}} \\
\midrule
\rowcolor{light-yellow} \multicolumn{3}{c|}{Shannon Entropy}        & 0.886  & \textbf{0.934}  & 0.773    \\
\rowcolor{mygray} 
\multicolumn{3}{c|}{Sentence Length}        & \textbf{0.890}  & {0.933}  & \textbf{0.776}    \\
\midrule
\multicolumn{6}{c}{{Ablation on the proposed adapter}} \\
\midrule
\rowcolor{rouse} \multicolumn{3}{c|}{Single Prompt}        & 0.705  & 0.920   & 0.765      \\
\rowcolor{rouse} \multicolumn{3}{c|}{Antonym Prompt}        & 0.875  & 0.928  & 0.771    \\
\rowcolor{mygray} \multicolumn{3}{c|}{Ours adapter}      & \textbf{0.890}  & \textbf{0.933} & \textbf{0.776}     \\
\midrule
\multicolumn{6}{c}{{Ablation on different MLLMs}} \\
\midrule
\rowcolor{mygray}  \multicolumn{3}{c|}{LLaVA-v1.5-7B}        & 0.871  & 0.914  & 0.755    \\
\rowcolor{light-blue} \multicolumn{3}{c|}{LLaVA-v1.5-13B}        & 0.872  & 0.914  & 0.757    \\
\rowcolor{light-blue} 
\multicolumn{3}{c|}{Sphinx}        & 0.874  & 0.916  & 0.758    \\
\rowcolor{light-blue} 
\multicolumn{3}{c|}{QWen-VL}        & 0.870  & 0.913  & 0.757    \\
\rowcolor{mygray} 
\multicolumn{3}{c|}{LLaVA-7B+QWen}        & {0.875}  & {0.916}  & {0.758}    \\
\rowcolor{mygray} 
\multicolumn{3}{c|}{Sphinx+QWen}        & \textbf{0.877}  & \textbf{0.918}  & \textbf{0.759}    \\
    \bottomrule
    \end{tabular}
}
    \label{tab:ablation}
\end{table}


\noindent \textbf{Zero-shot image retrieval.} We use different text queries to calculate the image-text similarity and rank them for zero-shot image retrieval. 
Fig.~\ref{fig:retri} shows the top retrieval results. 
We notice that ``good image'' prompts retrieve sharp, aesthetically pleasing images, whereas ``bad image'' prompts retrieve blurry, poor lighting and meaningless images. 
These examples provide qualitative evidence of the quality and aesthetic knowledge captured by the pre-trained model.

\noindent \textbf{More downstream datasets.} Our UniQA can be an effective foundation model for various downstream datasets. We evaluate on three datasets, \textit{i.e.}, AI-generated IQA dataset AIGC-3K~\cite{li2023agiqa3k} (Table~\ref{tab:aigc3k}), underwater IQA dataset UWIQA~\cite{uwiqa} (Table~\ref{tab:uwiqa}) and medical IQA dataset ECIQAD~\cite{yue2023perceptual} (Table~\ref{tab:eciqad}). We can observe that our UniQA can achieve SOTA performance on these three datasets.
Since images on these three datasets come from completely different scenarios, it is very challenging to consistently achieve the leading performance on
all of them. Correspondingly, these observations fully confirm that our UniQA has powerful generalization ability and can be used as an effective baseline method for various IQA scenarios.

\subsection{Data-Efficient Learning}

The pre-trained model acquires extensive image assessment knowledge, providing robust priors for downstream tasks. Consequently, our model can deliver impressive performance with limited data. To validate this, we randomly select subsets of 50, 100, and 200 samples from the training set for training and evaluate them on the same test data as full-data supervised learning. We report the median performance across 10 times in Table~\ref{tab:few-shot-sota}. Our method notably outperforms the second-best model GRepQ by a substantial margin, even though GRepQ is specifically designed for data-efficient learning. These results thoroughly demonstrate the potent capability of our method to learn image quality in few-label setting.
Additionally, several insightful observations can be drawn from Table~\ref{tab:few-shot-sota}. 
Firstly, the prompt ensemble strategy significantly boosts model performance in data-efficient settings by fully utilizing pre-trained model knowledge. Secondly, its impact on synthetic datasets is slight, likely due to limited image variety of synthetic images, making a single prompt adequate.

\subsection{Ablation Studies}

\noindent \textbf{Impact of different pre-training data.} 
Table~\ref{tab:ablation} explores the impact of different pre-training data.
We observe that unified pre-training achieves the optimal performance on both tasks. 
Additionally, we derive some meaningful observations.
(1) Using generated $Y_{IQA}$ or $Y_{IAA}$ improves the performance of both IQA and IAA tasks, proving the mutual benefit of two tasks and the effectiveness of MLLMs captioning. 
(2) Unifying $Y_{IQA}$ and $Y_{IAA}$ datasets does not lead to significant improvements. We believe this is because the MLLMs-generated text tends to have similar sentence structures~\cite{liu2023mllms} and representations, limiting the diversity provided for multimodal learning. (3) Pre-training with refined human-annotated $Y_{IAA}^{+}$ shows significant improvement on two tasks, indicating its comprehensive and effective representation for the model.

\begin{figure}[t]
\centering
\includegraphics[scale=0.34]{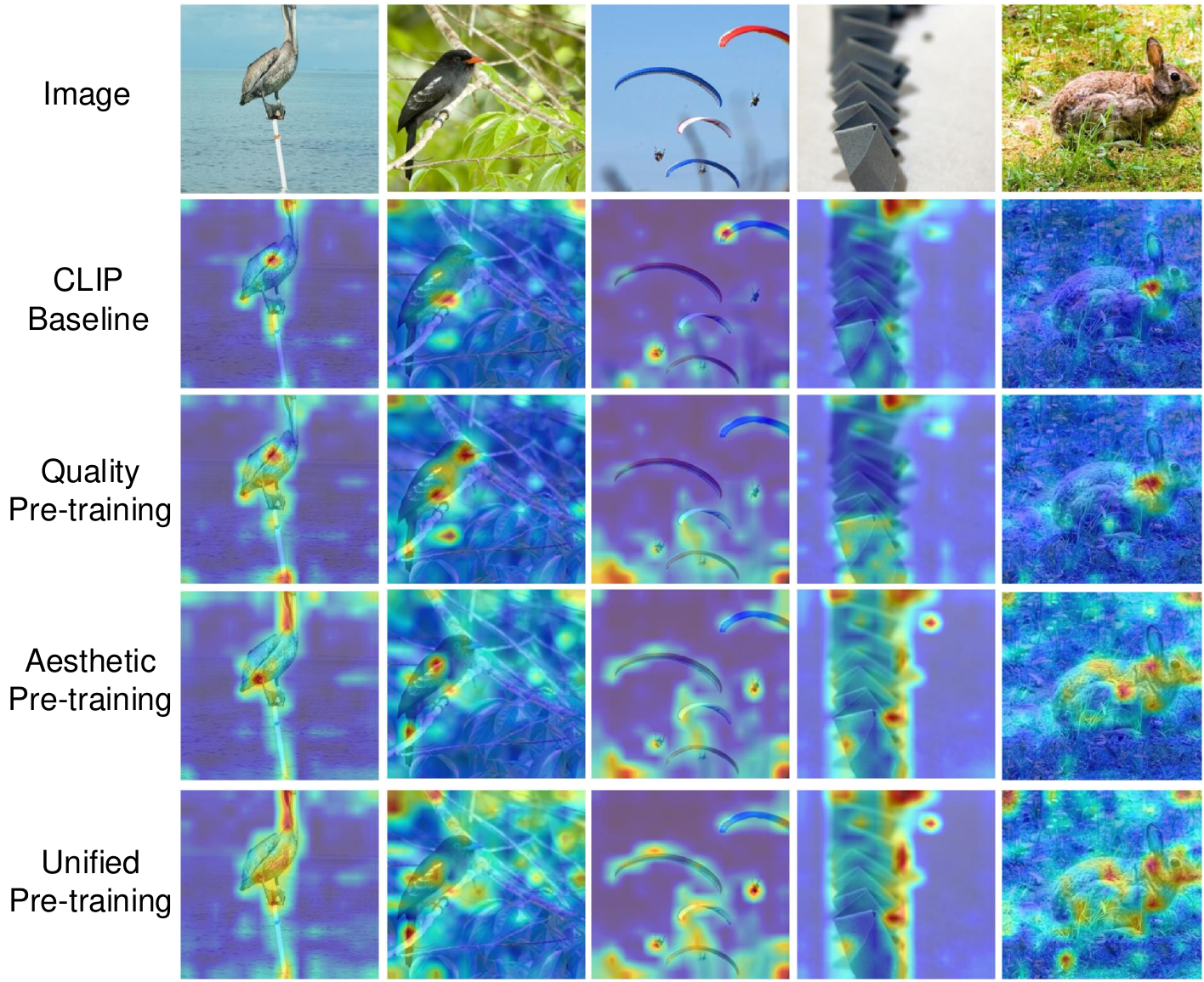}
\caption{ {Grad-CAM~\cite{selvaraju2017grad} visualization of different pre-training for prompt ``blurry image''. 
Through pre-training, the model focuses more on noisy objects and backgrounds.
}}
\label{fig:cam}
\end{figure}

Fig.~\ref{fig:cam} illustrates the Grad-CAM visualization of different pre-training. We notice that after quality  and aesthetic pre-training, the model pays more attention to blurred subjects and noisy backgrounds. 
This effect becomes more pronounced with unified pre-training, underscoring the advantages of such a unified approach. 
 {In addition, the unified pre-training can focus on the areas of quality and aesthetic pre-training at the same time. This shows that unified training can  learn common representations of the two tasks.}

\noindent 
\textbf{Effectiveness of data purification strategy.}
The second part of Table~\ref{tab:ablation} illustrates the ablation study of the data purification strategy. It can be observed that employing either AR or IR strategy to purify data can improve the model's performance  {of both IQA and IAA tasks.} 
These results validate the benefit of obtaining aesthetically relevant and informative descriptions for the model.
Finally, when combining these two strategies, the best performance is achieved.

\noindent 
\textbf{Discussion about the informativeness rank.} 
The third part of Table~\ref{tab:ablation} explores the effectiveness of different informativeness rank (IR) strategies. We can observe that using other informative methods (\textit{e.g.}, Shannon Entropy) yields similar performance to our IR. This is because, in addition to IR, we also use aesthetic relevance metrics to reflect the content quality of the text. Thus, even using simple sentence length as the informativeness measure can achieve satisfactory results.


\noindent 
\textbf{Effectiveness of the Multi-Cue Integration Adapter.}
The fourth part of Table~\ref{tab:ablation} evaluates the proposed adapter. ``Single Prompt'' denotes using the similarity between the text ``good image'' and images as the  score directly, while ``Antonym Prompt'' represents using the relative weights of texts ``good image'' and ``bad image'' to weight the predefined score. It is evident that the ``Single Prompt'' is considerably inferior to the ``Antonym Prompt'', showing the limitations of using semantic similarity as score directly. Our method  integrates more cues into the ``Antonym Prompt'' to comprehensively assess images, thereby achieving optimal performance.

\noindent 
\textbf{Ablation on different MLLMs.}
The bottom part of Table~\ref{tab:ablation} presents the ablation study of various MLLMs. We generate $Y_{IQA}$ via different MLLMs for pre-training. It is evident that using different MLLMs exhibits similar performance, while ensembling different MLLMs can boost performance. This indicates that MLLMs are capable of generating accurate captions with our text-guided prompt, and enhancing caption diversity can further improve performance. Considering resource limitations, we use LLaVA-7B and will integrate more MLLMs in the future.

\section{Conclusion}

This paper introduces UniQA, which leverages unified vision-language pre-training to address quality and aesthetic assessment problems concurrently.
We construct a high-quality image-text dataset about quality and aesthetics via MLLMs. Through large-scale pre-training on this dataset, UniQA learns shared and effective representations of IQA and IAA tasks,
enhancing the performance of two tasks significantly. 
In addition, we propose a Multi-Cue Integration Adapter to effectively adapt the pre-trained UniQA to downstream assessment tasks. Our method achieves state-of-the-art performance on both IQA and IAA tasks, and demonstrates powerful zero-shot and few-label image assessment capabilities.

\noindent \textbf{Limitations and future work.}  
MLLMs often generate captions with similar sentence structures and semantic expressions, restricting their ability to provide diverse and enriched representations for multimodal learning. 
Future work will explore other techniques to address this issue, including integrating various MLLMs for captioning and employing in-context learning methods.


\nocite{zhou2024unihead,zhou2023etdnet,xiao2023semanticac,chu2025attention,tang2024mind,zhou2024video}

\bibliographystyle{IEEEtran}
\bibliography{ref}


 





\end{document}